# Spatiotemporal information conversion machine for time-series prediction

Hao Peng[1], Pei Chen[1*], Rui Liu[1,2,*], Luonan Chen[3,4,5,6,*]

[1] School of Mathematics, South China University of Technology, Guangzhou 510640, China.

[2] Pazhou Lab, Guangzhou 510330, China.

[3] Key Laboratory of Systems Biology, Shanghai Institute of Biochemistry and Cell Biology, Center for Excellence in Molecular Cell Science, Chinese Academy of Sciences, Shanghai 200031, China.

[4] Center for Excellence in Animal Evolution and Genetics, Chinese Academy of Sciences, Kunming 650223, China.

[5] Key Laboratory of Systems Health Science of Zhejiang Province, Hangzhou Institute for Advanced Study, University of Chinese Academy of Sciences, Chinese Academy of Sciences, Hangzhou 310024, China.

[6] School of Life Science and Technology, ShanghaiTech University, Shanghai 201210, China.

* Correspondence: Luonan Chen, lnchen@sibs.ac.cn; Rui Liu, scliurui@scut.edu.cn; Pei Chen, chenpei@scut.edu.cn

**Abstract**

Making predictions in a robust way is a difficult task only based on the observed data of a nonlinear system. In this work, a neural network computing framework, the spatiotemporal information conversion machine (STICM), was developed to efficiently and accurately render a multistep-ahead prediction of a time series by employing a spatial-temporal information (STI) transformation. STICM combines the advantages of both the STI equation and the temporal convolutional network, which maps the high-dimensional/spatial data to the future temporal values of a target variable, thus naturally providing the prediction of the target variable. From the observed variables, the STICM also infers the causal factors of the target variable in the sense of Granger causality, which are in turn selected as effective spatial information to improve the prediction robustness of time-series. The STICM was successfully applied to both benchmark systems and real-world datasets, all of which show superior and robust performance in multistep-ahead prediction, even when the data were perturbed by noise. From both theoretical and computational viewpoints, the STICM has great potential in practical applications in artificial intelligence (AI) or as a model-free method based only on the observed data, and also opens a new way to explore the observed high-dimensional data in a dynamical manner for machine learning.

## 1. Introduction

It is difficult to render multistep-ahead predictions of a nonlinear dynamical system based on time-series data due to its complicated nonlinearity and insufficient information regarding future dynamics. Actually, great efforts have been devoted to solve this challenging problem. A number of methods including statistical regression (e.g., autoregressive integrated moving average (ARIMA) [1], robust regression [2]), exponential smoothing [3,4], and machine learning (e.g., long-short-term-memory (LSTM) network [5,6]), were utilized in forecasting unknown states [7–10]. However, most of them cannot make satisfactory predictions regarding short-term time series due to insufficient information. To solve this problem, the auto-reservoir neural network (ARNN) [11]) was developed by using the semi-linearized spatial-temporal information (STI) transformation equation [11,12], which transforms high-dimensional information into temporal dynamics of any target variable, thus effectively extending the data size. However, this approach does not fully explore the nonlinearity of the STI equation from the observed data, which is essential for accurately predicting many complex systems. In addition, few existing approaches take spatial and temporal causal interactions of high-dimensional time-series data into consideration, which can compensate for insufficient data and provide reliable information to predict a complex dynamical system.

Under the condition that the steady state of a high-dimensional dynamical system is contained in a low-dimensional manifold, which is actually satisfied for most real-world systems, the STI transformation equation has theoretically been derived from delay embedding theory [13–15]. This equation can transform the spatial information of high-dimensional data into the temporal information of any target variable, thus equivalently expanding the sample size. Based on the STI transformation, the randomly distributed embedding (RDE) method was proposed to predict the one-step-ahead value from high-dimensional time-course data by separately constructing multiple STI maps (or primary STI equations) to form the distribution of the predicted values [12]. Our recent auto-reservoir computing framework ARNN [11] achieves multistep-ahead prediction based on a semi-linearized STI transformation; however, the nonlinear features and spatiotemporal causal relations of the observed high-dimensional variables have not yet been exploited, which restricts the prediction performance in the sense of robustness and accuracy.

On the other hand, a temporal convolutional network (TCN) [16] was recently reported to outperform canonical recurrent neural networks (RNNs) [17–19], such as the LSTM network [5,6], and the gated recurrent units (GRU)[20], across a diverse range of sequence modelling tasks and datasets. Compared with RNNs, the TCN possesses advantages including a longer effective memory length, a flexible receptive field size, stable gradients, a low memory requirement for training, variable-length inputs, and parallelism [16]. Besides, the TCN employs dilated convolution, which enables an exponentially large receptive field, to handle long sequences. However, the traditional TCN does not fully reveal the causal relations among high-dimensional variables and cannot make accurate multi-step ahead prediction without future information/labels.

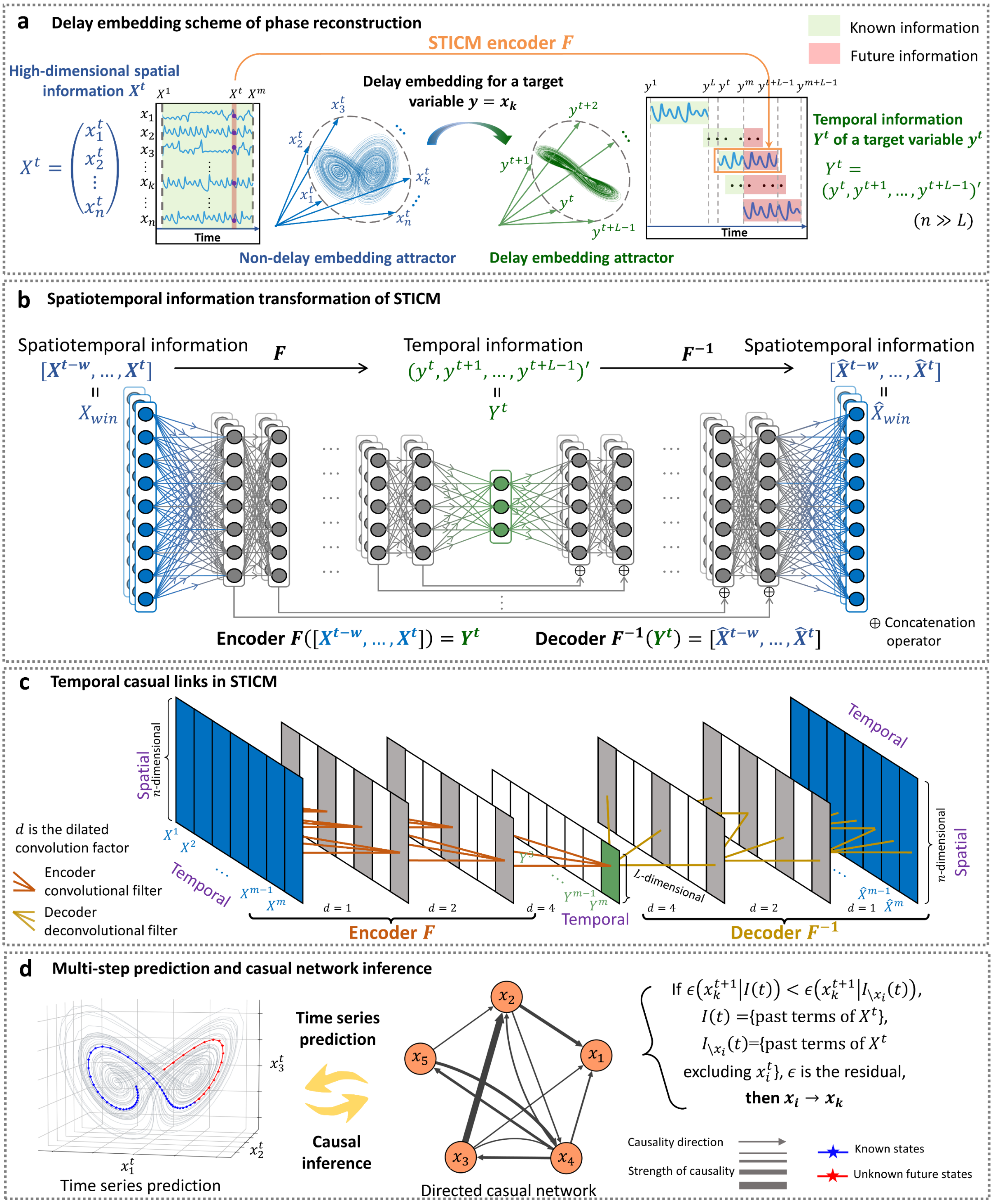

Fig. 1 Schematic illustration of the spatiotemporal information conversion machine (STICM). (a) For a to-be-predicted/target variable $y$ selected from the high-dimensional observables $\{x_1, x_2, \ldots, x_n\}$, a temporal vector $\boldsymbol{Y}^t$ is constructed through a delay embedding scheme. The temporal vector $\boldsymbol{Y}^t$ is corresponding to an observed spatiotemporal matrix $[\boldsymbol{X}^{t-w}, \boldsymbol{X}^{t-w+1}, \ldots, \boldsymbol{X}^t]$ via a nonlinear function $F$. (b) The information flow of the STICM is similar to the autoencoder (AE) but is constrained by primary and conjugate STI equations. The primary STI equation represents the encoder, while the conjugate STI equation corresponds to the decoder. However, unlike AE, the low-dimensional/temporal code $Y^t$ is mapped by the delay embedding scheme from the time series of a target variable $y$. (c) The encoder and decoder of STICM are implemented by a temporal convolutional network (TCN) structure and a temporal deconvolutional structure, respectively, through which the spatiotemporal matrix

$[\boldsymbol{X}^1, \boldsymbol{X}^2, \dots, \boldsymbol{X}^t]$ is input sequentially and mapped to $[\boldsymbol{Y}^1, \boldsymbol{Y}^2, \dots, \boldsymbol{Y}^t]$. (d) By inferring the causal relations and selecting the effective variables, the prediction performance is considerably improved. Note that the mapping $F$ is from a matrix $[\boldsymbol{X}^{t-w}, \boldsymbol{X}^{t-w+1}, \dots, \boldsymbol{X}^t]$ to a vector $\boldsymbol{Y}^t$.

In this study, we propose a novel framework, i.e., spatiotemporal information conversion machine (STICM), to achieve accurate and robust multistep-ahead prediction with high-dimensional data, and explore the underlying causal relations among high-dimensional variables. The central idea is to represent both primary and conjugate STI equations in an autoencoder form (Fig. 1) by exploiting the advantages of the causal convolution and STI nonlinear transformation. Computationally, the STICM includes three basic processes: 1) the embedding scheme to reconstruct the phase space (Fig. 1a), 2) the STICM to realize the STI transformation (Fig. 1b, c), and 3) effective/causal variable selection to make the prediction more accurate and robust (Fig. 1d). In particular, we adopt both the primary and the conjugate forms of the STI equations to encode (through nonlinear function $F$) and decode (through the reverse function $F^{-1}$) the temporal dynamics from the high-dimensional data (see Fig. 1b and Eq. (18)). Through the STI equations, the STICM transforms the spatiotemporal information of high-dimensional data to the temporal/dynamical future values of a target variable. Given the time-course data of high-dimensional variables, the STICM trains the encoder $F$ and decoder $F^{-1}$ by taking both spatial and temporal information into consideration (Fig. 1b, c), thus equivalently expanding the data size on the target variable or naturally resulting in the future values of the target variable $y$. Moreover, by comparing the prediction error, the STICM directly makes the Granger inference of causal factors on the target variable, which are in turn selected as the effective/spatial variables to significantly improve the prediction robustness and accuracy of the target variable.

To validate the accuracy and robustness, STICM was applied to a series of representative mathematical models, i.e., a 90-dimensional coupled Lorenz system [21] under different noise conditions. Furthermore, the STICM was applied to many real-world datasets in this study and predicted, e.g., 1) the daily number of cardiovascular inpatients in the major hospitals of Hong Kong [22,23], 2) the wind speed in Japan [24], 3) a ground meteorological dataset in the Houston, Galveston, and Brazoria areas [25], 4) the population of the plankton community isolated from the Baltic Sea [26,27], 5) the spread of COVID-19 in the Kanto region of Japan [28], and 6) the traffic speed of multiple locations in Los Angeles [29]. The results show that the STICM achieves multistep-ahead prediction that is better than the other seven existing methods in terms of accuracy and robustness. More descriptions of each compared method are illustrated in Supplementary Section 6. As a model-free method based only on the observed data, the STICM framework paves a new way to make multistep-ahead predictions by incorporating the primary-conjugate STI equations into an autoencoder form. This framework exploits both the STI transformation and causal convolutional structure, thus is of great potential for practical applications in many scientific and engineering fields, and also opens a new way to dynamically explore high-dimensional information in machine learning.

## 2. Methods

The detailed description of the parameters and variables in STICM framework are summarized in Supplementary Table S3.

### 2.1. Delay embedding theorem for dynamical systems

Generally, the dynamics of a discrete-time dissipative system can be presented as

$$\mathbf{X}^{t+1} = \phi(\mathbf{X}^t), \tag{1}$$

where $\phi: \mathbb{R}^n \to \mathbb{R}^n$ represents a nonlinear function, whose $n$-dimensional variables are denoted as vector $\mathbf{X}^t = (x_1^t, x_2^t, \dots, x_n^t)'$ with time superscript $t$ and vector transpose symbol " $'$ ". The Takens' embedding theorem provides the following facts [13,14].

If $\mathcal{V} \subseteq \mathbb{R}^n$ is a compact attractor with the Minkowski dimension/box-counting dimension $d$, for a smooth diffeomorphism $\phi: \mathcal{V} \to \mathcal{V}$ and a smooth function $h: \mathcal{V} \to \mathbb{R}$, there is a generic property that the mapping $\Phi_{\phi,h}: \mathcal{V} \to \mathbb{R}^L$ is an embedding when $L > 2d$, that is,

$$\Phi_{\phi,h}(X) = \left(h(X), h \circ \phi(X), \dots, h \circ \phi^{L-1}(X)\right)', \tag{2}$$

where symbol " $\circ$ " is the function composition operation. In particular, letting $X = \mathbf{X}^t$ and $h(\mathbf{X}^t) = y^t$ where $y^t \in \mathbb{R}$, then the mapping above has the following form with $\Phi_{\phi,h} = \Phi$ and

$$\Phi(\mathbf{X}^t) = (y^t, y^{t+1}, \dots, y^{t+L-1})' = \mathbf{Y}^t. \tag{3}$$

Moreover, since the embedding is one-to-one mapping, we can also derive its conjugate form $\Psi: \mathbb{R}^L \to \mathbb{R}^n$ as $\mathbf{X}^t = \Phi^{-1}(\mathbf{Y}^t) = \Psi(\mathbf{Y}^t)$ (Supplementary Section 1). Here $\mathbf{X}^t$ is $n$- dimensional variables here, but sometimes it is used as $D$-dimensional variables ($D \leq n$) in this work. The above theory can be summarized as the following spatiotemporal information (STI) transformation equation:

$$\begin{cases} \Phi(\mathbf{X}^t) = \mathbf{Y}^t, \\ \mathbf{X}^t = \Psi(\mathbf{Y}^t), \end{cases} \tag{4}$$

where $\Phi: \mathbb{R}^D \to \mathbb{R}^L$ and $\Psi: \mathbb{R}^L \to \mathbb{R}^D$ are nonlinear differentiable functions satisfying $\Phi \circ \Psi = id$, symbol " $\circ$ " represents the function composition operation and $id$ denotes the identity function.

Note that to use the causal convolution framework of TCN, we let $X = [\mathbf{X}^{t-w}, \mathbf{X}^{t-w+1}, \dots, \mathbf{X}^t]$ in Eq. (3) for this work with map $F$ (*i.e.*, Eq. (18)), rather than $X = \mathbf{X}^t$ with map $\Phi$.

## 2.2. STICM algorithm

The determination of $F$ and $F^{-1}$ includes two main factors: 1) the semi-supervised training scheme and 2) the effective variable selection. This structure of STICM is capable of exploiting not only the input of spatial information but also the temporally intertwined information among the numerous variables of the complex dynamic system, thus greatly enhancing the forecasting robustness and accuracy. In this study, each layer of the encoder $F$ and decoder $F^{-1}$ is followed by the ReLU activation function. The STICM algorithm is carried out to uncover the to-be-predicted/future values $\{y^{m+1}, y^{m+2}, \dots, y^{m+L-1}\}$ of the target $y = x_k$ with the following procedure.

**Step 1: Construct the STICM-based STI equation.** Based on the delay embedding scheme, we construct the delay-embedded matrix of the target variable $y$ as Eq. (17) with the columns $\mathbf{Y}^t = (y^t, y^{t+1}, \dots, y^{t+L-1})'$. Clearly, vectors $\{\mathbf{Y}^{m-L+2}, \dots, \mathbf{Y}^m\}$ contains the unknown/future values. The steady state or the attractor is generally constrained in a low-dimensional space for a high-dimensional dissipative system, which holds for most real-world systems. Assuming $F = (F_1, F_2, \dots, F_L)'$ and $L > 2d$ where $d$ is the Minkowski dimension of the attractor, the primary form of the STICM-based STI equation set (Eq. (18)) is

$$\begin{bmatrix} F_1([\mathbf{X}^1]) & F_1([\mathbf{X}^1,\mathbf{X}^2]) & \cdots & F_1([\mathbf{X}^{m-w},\dots,\mathbf{X}^m]) \\ F_2([\mathbf{X}^1]) & F_2([\mathbf{X}^1,\mathbf{X}^2]) & \cdots & F_2([\mathbf{X}^{m-w},\dots,\mathbf{X}^m]) \\ \vdots & \vdots & \ddots & \vdots \\ F_L([\mathbf{X}^1]) & F_L([\mathbf{X}^1,\mathbf{X}^2]) & \cdots & F_L([\mathbf{X}^{m-w},\dots,\mathbf{X}^m]) \end{bmatrix} = \begin{bmatrix} y^1 & y^2 & \cdots & y^m \\ y^2 & y^3 & \cdots & y^{m+1} \\ \vdots & \vdots & \ddots & \vdots \\ y^L & y^{L+1} & \cdots & y^{m+L-1} \end{bmatrix}. \tag{5}$$

In a similar form of Eq. (5), we have the conjugate equation with $F^{-1}$ (see Supplementary Eq. (7)). Clearly, by simultaneously solving both the primary and conjugate STICM-based STI equations, the STICM provides a series of future values $\{y^{m+1}, y^{m+2}, \dots, y^{m+L-1}\}$, which is indeed the $(L-1)$-step-ahead prediction.

**Step 2: Train the STICM network.** Because there are both known and unknown values in the delay embedding matrix $Y$, the STICM is trained in a semi-supervised manner. Specifically, the nonlinear mappings $F = (F_1, F_2, \dots, F_L)'$ are fit via a "consistently self-constrained scheme" simultaneously for preserving the time consistency for the known and unknown values, thus maintaining the integrity of $F$. According to the framework of STICM, there are three high-level requirements for the network used in training.

Due to the delay-embedding nature in the output $Y$ (as shown in Eq. (5)), we have totally $m + L - 3$ temporally self-constrained conditions

$$\begin{aligned} &F_{j-1}([\mathbf{X}^{t-w}, \mathbf{X}^{t-w+1}, \dots, \mathbf{X}^t]) = \\ &F_j([\mathbf{X}^{t-w-1}, \mathbf{X}^{t-w}, \dots, \mathbf{X}^{t-1}]), \end{aligned} \tag{6}$$

where $j \in \{2, 3, \dots, L\}$ and $\mathbf{X}^t = (x_1^t, x_2^t, \dots, x_n^t)'$ is a spatial sample at time point $t$. Among conditions Eq. (6), there are $m-1$ conditions for the determined states and $L-2$ conditions for future values. Clearly, these conditions constrain the training of STICM in terms of the temporal sequence of samples. For the target variable $y$, the estimated values of its delay embeddings in each iteration are obtained as follows

$$\hat{Y} = \begin{bmatrix} (\hat{y}^1)_1 & (\hat{y}^2)_1 & \cdots & (\hat{y}^m)_1 \\ (\hat{y}^2)_2 & (\hat{y}^3)_2 & \cdots & (\hat{y}^{m+1})_2 \\ \vdots & \vdots & \ddots & \vdots \\ (\hat{y}^L)_L & (\hat{y}^{L+1})_L & \cdots & (\hat{y}^{m+L-1})_L \end{bmatrix}, \tag{7}$$

where $(\hat{y}^t)_j$ ($t = 1, 2, \dots, m+L-1$; $j = 1, 2, \dots, L$) is generated from the output of the $j^{\text{th}}$ sub-mapping function $\mathcal{F}_j$.

Through an auto perception procedure, the training or optimization of STICM is accomplished through a process of minimizing a loss function with three weighted mean-squared error components

$$\mathcal{L} = \lambda_1 \mathcal{L}_{DS} + \lambda_2 \mathcal{L}_{FC} + \lambda_3 \mathcal{L}_{REC}. \tag{8}$$

In Eq. (8), the first part $\mathcal{L}_{DS}$ is a determined-state loss from the observed/known states $\{y^1, y^2, \dots, y^m\}$ of $y$, and is of the following form

$$\mathcal{L}_{DS} = \frac{1}{2mL - L^2 + L} \sum_{j=1}^{L} \sum_{t=j}^{m} \left((\hat{y}^t)_j - y^t\right)^2, \tag{9}$$

where $(\hat{y}^t)_j$ $(t = 1, 2, \dots, m)$ is the estimation of $\mathcal{F}_j([\mathbf{X}^{t-w}, \mathbf{X}^{t-w+1}, \dots, \mathbf{X}^t])$, and $y^t$ $(t = 1, 2, \dots, m)$ is the known value of $y$. Loss $\mathcal{L}_{DS}$ is constructed from the differences between the estimations $(\hat{y}^t)_j$ and the observed values (ground truth) $y^t$ for all past time points $t$ $(t = 1, 2, \dots, m)$.

In Eq. (8), the second part $\mathcal{L}_{FC}$ is a future-consistency loss in terms of the future/unknown series $\{y^{m+1}, y^{m+2}, \dots, y^{m+L-1}\}$ of $y$, and has form

$$\mathcal{L}_{FC} = \frac{1}{L(L-1)} \sum_{j=2}^{L} \sum_{t=m+1}^{m+j-1} \left((\hat{y}^t)_j - \mathrm{mean}(\hat{y}^t)\right)^2, \tag{10}$$

where $\mathrm{mean}(\hat{y}^t)$ denotes the average of all estimated future values of $\hat{y}^t$ in Eq. (7) that corresponds to the same future time point $t$ $(t = m+1, m+2, \dots, m+L-1)$. Clearly, $\mathcal{L}_{FC}$ is constructed from the temporally self-constrained conditions in Eq. (6) An intuitive understanding of the future-consistency loss is that by minimizing $\mathcal{L}_{FC}$, it ensures that the outputs from different sub-mappings but corresponding to the same future time point $t$ are identical, which preserves the temporal consistency of the outputs at the lower right corner of the delay embedding matrix $\hat{Y}$ during the training procedure.

In Eq. (8), the third part $\mathcal{L}_{rec}$ is a reconstruction loss in terms of the consistency of encoder and decoder, which is of the following form

$$\mathcal{L}_{rec} = \| X - \hat{X} \|_{\mathrm{F}}, \tag{11}$$

where $X = [\mathbf{X}^1, \mathbf{X}^2, \dots, \mathbf{X}^t]$, $\hat{X} = [\hat{\mathbf{X}}^1, \hat{\mathbf{X}}^2, \dots, \hat{\mathbf{X}}^t]$, and $\|\cdot\|_{\mathrm{F}}$ is the Frobenius norm.

Based on the integration of the above three losses, STICM is trained in a semi-supervised manner. The cooperation of future-consistency loss $\mathcal{L}_{FC}$ and determined-state loss $\mathcal{L}_{DS}$ helps to fit the nonlinear mapping $F = (F_1, F_2, \dots, F_L)'$. The reconstruction loss $\mathcal{L}_{rec}$ guarantees the consistency of encoder and decoder. After the convergence of the training process, the $m+L-1$ to-be-predicted values $\{y^{m+1}, y^{m+2}, \dots, y^{m+L-1}\}$ can eventually be determined from the estimated matrix $\hat{Y}$, *i.e.*,

$$y^{m+i} = \text{mean}\left(\hat{y}^{m+i}\right) = \frac{1}{L-i}\sum_{j=i+1}^{L}\left(\hat{y}^{m+i}\right)_j, \tag{12}$$

with $i = 1, 2, \dots, L-1$. The implementation of the deconvolution layer in decoder $F^{-1}$ is similar and provided in Supplementary Section 4 and Fig. S1c.

**Step 3: Identify the causal/driving variables.**

To decrease the noisy effect and boost the robustness on the prediction, we choose the most relevant variables to the target variable from the high-dimensional data. Given a time series of $n$-dimensional samples $(x_1^t, x_2^t, \dots, x_n^t)'_{t=1,2,\dots,m}$, we calculate the prediction errors between the case "with an observable $x_i$" and the case "without $x_i$". Then, one can determine whether $x_i$ is a causal/effective factor of the target variable $y$ in the sense of Granger causality, thus improve the prediction by selection or deletion of the variable.

First, a reference RMSE $\epsilon_r$ of the model trained by the original $n$-dimensional input was calculated as the normalized difference between original and predicted values, *i.e.*,

$$\epsilon_r = \text{RMSE}(y, \hat{y}|\Lambda) = \sqrt{\frac{\sum_{t=m}^{m+L-1}(y^t - \hat{y}^t|\Lambda^t)^2}{L-1}}, \tag{13}$$

where $y^t$ denotes the original value of the target variable, $\hat{y}^t$ denotes the predicted one, and $\Lambda^t$ denotes the past terms of $X^t$. Subsequently, by excluding $x_i, i = 1,2,\dots,n$ from the original data, the model is trained based on an $(n-1)$-dimensional input with a test RMSE $\epsilon_i$,

$$\epsilon_i = \text{RMSE}(y, \hat{y}|\Lambda\backslash x_i) = \sqrt{\frac{\sum_{t=m}^{m+L-1}(y^t - \hat{y}^t|\Lambda^t\backslash x_i^t)^2}{L-1}}, \tag{14}$$

where $\Lambda^t\backslash x_i^t$ denotes the past terms of $X^t$ without $x_i^t$. Then, a causality error $\epsilon_{i,r}$ is obtained as

$$\epsilon_{i,r} = \epsilon_i - \epsilon_r, \tag{15}$$

which denotes the influence of Granger causality from variable $x_i$ to $y$. After ranking all $\epsilon_{i,c}$ $(i = 1,2,\dots,n)$, we selected the spatial information of top $q$ causal/effective variables as new input data, which are most relevant to the target variable $y$. By excluding irrelevant variables or noisy information, the final STICM is trained based on such a lower-dimensional input and enhances the prediction performance in terms of both accuracy and robustness. The schematic illustration of this step can be found in Fig. S1a. The other details of the STICM algorithm are provided in Supplementary Section 3.

## 3. Results

### 3.1. STICM framework with STI transformation

For each observed high-dimensional/spatial state $\mathbf{X}^t = (x_1^t, x_2^t, \ldots, x_n^t)'$ with $n$ variables with $t = 1,2, \ldots, m$, a corresponding delayed/temporal vector $\mathbf{Y}^t = (y^t, y^{t+1}, \ldots, y^{t+L-1})'$ is constructed for one target variable $y$ (e.g., $y^t = x_k^t$) through a delay embedding scheme with parameter $L$ as the embedding dimension satisfying $n > L > 1$ (Fig. 1a), where symbol " $'$ " is the transpose of a vector. Specifically, the matrix $X$ of the original measurable variables $\{x_1, x_2, \ldots, x_n\}$ is as follows:

$$X = [\mathbf{X}^1, \mathbf{X}^2, \ldots, \mathbf{X}^m] = \begin{bmatrix} x_1^1 & x_1^2 & \cdots & x_1^m \\ x_2^1 & x_2^2 & \cdots & x_2^m \\ \vdots & \vdots & \ddots & \vdots \\ x_n^1 & x_n^2 & \cdots & x_n^m \end{bmatrix}_{D\times m} . \tag{16}$$

Through the delay embedding scheme, the matrix $Y$ of the target variable $y = x_k$ is

$$Y = \begin{bmatrix} y^1 & y^2 & \cdots & y^m \\ y^2 & y^3 & \cdots & y^{m+1} \\ \vdots & \vdots & \ddots & \vdots \\ y^L & y^{L+1} & \cdots & y^{m+L-1} \end{bmatrix}_{L\times m} , \tag{17}$$

where $Y$ contains the unknown/future values $\{y^{m+1}, y^{m+2}, \ldots, y^{m+L-1}\}$ in the lower-right corner (shadow area) of the target variable. It is clear that $\mathbf{X}^t$ is a known high-dimensional/spatial vector for multiple variables at one time point $t$, while $\mathbf{Y}^t$ is a temporal vector of one target $y$ at multiple time points $t$, $t + 1, \ldots,$ $t + L - 1$.

Based on the generalized Takens' embedding theory, the dynamics of the original system can be topologically reconstructed from a delay embedding scheme if $L > 2d > 0$, where $d$ is the Minkowski dimension of the attractor [13,14]. By combining the causal convolution structure and STI transformation, we developed an STCIM framework, which provides multistep-ahead prediction with dynamic causal inference among the observed variables on the basis of both the primary and conjugate STI equations (Fig. 1b). The known high-dimensional time series, *i.e.*, one sliding window matrix $X_{win} = [\mathbf{X}^{t-w}, \mathbf{X}^{t-w+1}, \ldots, \mathbf{X}^t]$ with window size $w+1$ from the whole spatiotemporal matrix $X = [\mathbf{X}^1, \mathbf{X}^2, \dots, \mathbf{X}^t]$, are mapped to one temporal delayed vector $\boldsymbol{Y}^t$ for $t = 1,2, \dots, m$, which actually forms the following STCIM -based STI equation set:

$$\begin{cases} F([\mathbf{X}^{t-w}, \mathbf{X}^{t-w+1}, \dots, \mathbf{X}^t]) = \mathbf{Y}^t, \\ F^{-1}(\mathbf{Y}^t) = [\widehat{\mathbf{X}}^{t-w}, \widehat{\mathbf{X}}^{t-w+1}, \dots, \widehat{\mathbf{X}}^t], \end{cases} \tag{18}$$

where the first formula is the primary equation with $F: \mathbb{R}^{n\times(w+1)} \to \mathbb{R}^L$ and the second formula is the conjugate equation with $F^{-1}: \mathbb{R}^L \to \mathbb{R}^{n\times(w+1)}$ (Fig. 1b), $\widehat{\mathbf{X}}^t$ is the recovered vector of $\mathbf{X}^t$. Given $m$ known states $\boldsymbol{X}^t$ $(t = 1,2, \dots, m)$, there are $L-1$ future values of $y$, *i.e.*, $\{y^{m+1}, y^{m+2}, \dots, y^{m+L-1}\}$ in $\boldsymbol{Y}^t$ (Fig. 1a and Fig. S1b). Matrix $[\mathbf{X}^{t-w}, \mathbf{X}^{t-w+1}, \dots, \mathbf{X}^t]$ of Eq. (18) is the known spatiotemporal information of $n$ variables, and $Y^t$ presents the temporal information of the target variable. In Eq. (18), the first and second equations are the primary and conjugate form of the STI equations, respectively. The primary equation encodes one spatiotemporal matrix $[\mathbf{X}^{t-w}, \mathbf{X}^{t-w+1}, \dots, \mathbf{X}^t]$ to one temporal vector $\mathbf{Y}^t$, while the conjugate form decodes/recovers the encoded temporal information $\mathbf{Y}^t$ to the spatiotemporal information $[\widehat{\mathbf{X}}^{t-w}, \widehat{\mathbf{X}}^{t-w+1}, \dots, \widehat{\mathbf{X}}^t]$. The STI equations (Eq. (18)) hold when some generic conditions are satisfied according to the delay embedding theory [13,14]. Clearly, the properly determined function $F$ is the key to solving the STCIM -based STI equations (Eq. (18)) for the high-dimensional input/matrix $X$ and providing the future values $\{y^{m+1}, y^{m+2}, \dots, y^{m+L-1}\}$ of the target variable. The details of Takens' embedding theory and the STI equations are given in Supplementary Section 1, and Supplementary Section 2, respectively.

The dilated causal convolution layers are employed in the framework of STCIM (Fig. 1c), that is, for input series $X = [\mathbf{X}^1, \mathbf{X}^2, \dots, \mathbf{X}^m]$ and a filter $g: \{0, \dots, k-1\} \to \mathbb{R}$, the dilated causal convolution operation $G$ on element $\mathbf{X}^t$ is defined as

$$G(\mathbf{X}^t) = (X *_d g)(\mathbf{X}^t) = \sum_{i=0}^{k-1} g(i) \cdot \mathbf{X}^{t-d\cdot i} \tag{19}$$

where $d$ is the dilation factor, $k$ is the filter size. Dilation is thus equivalent to introducing a fixed step between every two adjacent filter taps. A larger dilation enables an output at the top level to represent a wider range of inputs, thus effectively expanding the receptive field of a ConvNet. In this way, we construct the network structure for encoder $F$. Similarly, we adopted an inverse dilated convolution scheme in decoder $F^{-1}$, which is shown in Supplementary Section 4 in details.

## 3.2. Performance of the STICM on Lorenz models

To demonstrate the basic idea of the STCIM method, the synthetic time-series datasets under multiple noise levels were generated from a benchmark nonlinear system, *i.e.*, the following 90-dimensional coupled Lorenz model ($n = 90$) [21]

$$\dot{\mathbf{X}}(t) = G(\mathbf{X}(t); P) \tag{20}$$

where $P$ is a parameter vector of the function set $G(\cdot)$with $\mathbf{X}(t) = (x_1^t, x_2^t, \dots, x_{90}^t)'$. The specific Lorenz system is presented in Supplementary Section 5.

### 3.2.1. Noise-free situation

We first apply the STICM to a noise-free Lorenz system (Eq. (20)) with $m = 50$ and $L - 1 = 15$, *i.e.*, taking a time series of 50 steps as known information/input, and making a 15-step-ahead prediction/output for the target variables. As demonstrated in Fig. 2, the STICM predicted the future values for both the single-wing (Fig. 2c, the observed and to-be-predicted series distributed in a single wing of the attractor) and cross-wing (Fig. 2d, the observed and to-be-predicted series distributed in two different wings of the attractor) cases. By randomly selecting three target variables $y_1$, $y_2$ and $y_3$ from $\{x_1, x_2, \dots, x_{90}\}$, the prediction performances of the STICM on three-dimensional cases are presented in Fig. 2a and 2b. Notably, the predicted values (the red curves) for each target variable were obtained by the one-time prediction; that is, the STICM provides an efficient way to obtain a whole horizon (15 steps) of future information. Clearly, on the basis of the 90-dimensional short-term time series, the STICM inferred the top 30 effective/causal variables of the targets and significantly increased the performance in both accuracy and robustness by applying the prediction of the target with these 30 variables (Fig. 2c and 2d, Table 1 and Table S1). Note that the training and prediction of the STICM are based only on the observed data.

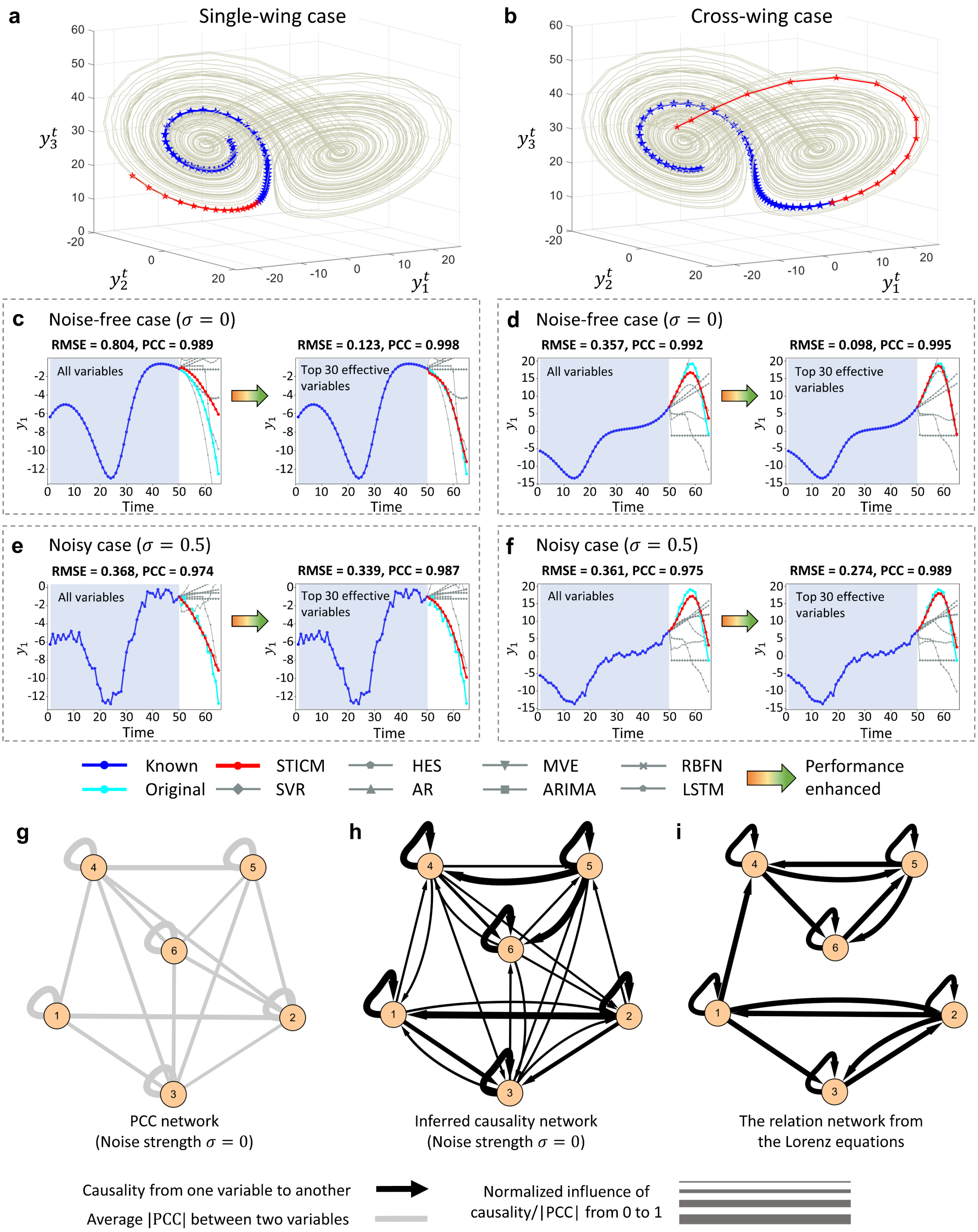

Fig. 2 The prediction performance of the STICM on the high-dimensional Lorenz system. In noise-free or noisy situations, the time-series data were generated on the basis of the 90D coupled Lorenz system (Eq. (20)). We randomly selected three targets $y_1$, $y_2$ and $y_3$ among variables $\{x_1, x_2, \ldots, x_{90}\}$. By applying the STICM with parameter $m = 50$ (*i.e.*, the length of the input series is 50), the 15-step-ahead predictions ($L - 1 = 15$) were performed for $y_1$, $y_2$ and $y_3$, respectively. (a) The prediction of the 3D system of $y_1$, $y_2$ and $y_3$ in the single-wing situation. (b) The prediction of the 3D system in the cross-wing situation. (c) The prediction of $y_1$ in a noise-free case of the single-wing situation. (d) The prediction of $y_1$ in a noise-free case of the cross-wing situation. (e) The prediction of target $y_1$ in a noisy case (with noise strength $\sigma = 0.5$) of the single-wing situation. (f) The prediction of $y_1$ in a noisy case (with noise strength $\sigma = 0.5$) of the cross-wing situation. For each case, the predictions are carried out based on all variables (the left panels of (b), (c), (d), and (e)) and based on the top 30 causal variables (the right panels of (b), (c), (d), and (e)). The PCC network (each edge is weighted with Pearson correlation coefficient) and causal relation network (each edge is weighted with Granger causality index) of the six selected effective variables in the noisy-free case (g) and (h), respectively. These two networks in the noisy case (with noise strength $\sigma = 0.5$) are illustrated in Fig. S4. (i) The relation network of the six variables from the Lorenz equations.

Here and below, to validate the effectiveness of the STICM, its prediction performance was compared with seven representative methods, *i.e.*, the LSTM network [5,6], Holt's exponential smoothing (HES) [3,4], autoregression (AR) [30], autoregressive integrated moving average (ARIMA) [1], radial basis function network (RBFN) [31], multiview embedding (MVE) [32], and support vector regression (SVR) [33,34]. Additionally, from Table 1 that the STICM achieved better performances compared with other prediction methods on the noise-free cases of the 90-dimensional Lorenz system; that is, the accuracy of the STICM is the best in terms of the Pearson correlation coefficient (PCC) and the root mean square error (RMSE). Specifically, the RMSEs decreased from 0.804 to 0.123 and from 0.357 to 0.098 for cases in Fig. 2c and Fig. 2d, respectively. As shown in Table 1, the STICM achieved the smallest RMSE 0.111 in the noise-free situation, while the best record of the other methods is 0.806. In addition, the inferred causality network among the six selected variables (Fig. 2h) is consistent with the direct causal relations from the original equations (Fig. 2g), and fully reveals intrinsic dynamic associations (including both direct and indirect causal relations) of the coupled Lorenz system comparing with the PCC network (Fig. 2g). Note that the direct causal relation from $x_i$ to $x_j$ in Fig. 2i is determined if $x_i$ is one of the bases/independent variables of $x_j$ in Supplementary Eq. (17). Moreover, the performances of eight prediction methods on the datasets without effective/causal variable selection are shown in Supplementary Table S1.

### 3.2.2. Additive noise situation

Second, the STICM was applied to the noisy cases of the 90D Lorenz system (Eq. (20)) with additive white noise ($\sigma = 0.5$) to predict the same target variable, while $m = 50$, and $L - 1 = 15$. Specifically, the cross-wing case is exhibited in Fig. 2f, and the single-wing case is presented in Fig. 2e. After the selection of the top 30 effective/causal variables, the prediction accuracy of the STICM improves significantly and is better than that of the other seven methods for both the single-wing and cross-wing cases (Table 1 and Table S1). Specifically, the RMSEs of our proposed method decreased from 0.368 to 0.339 (Fig. 2e) and from 0.361 to 0.274 (Fig. 2f). The average RMSE (0.307) of STICM across all noisy cases is the best among all prediction methods (Table 1 and Supplementary Table 1). Therefore, the STICM still predict the future dynamics accurately when the system is perturbed by additive noise, which demonstrates the robustness property of the STICM framework.

The STICM achieves satisfactory performance even with noisy data compared with traditional approaches because of its two characteristics, that is, simultaneously solving both conjugated STI equations in Eq. (18), and effective variable selection among all observables.

### 3.3. The application of the STICM on real-world datasets

Predicting the future values of key variables by exploiting the relevant high-dimensional information is of great importance for studying complex systems forecasting potential risk. The STICM method was applied to the following various high-dimensional real-world datasets, and was also compared with seven existing methods. The detailed performances of all the prediction methods are exhibited in Table 1. For each dataset, the specific settings and parameters are presented in Supplementary Table S2. The description of each dataset is also provided in Supplementary Section 5.

#### 3.3.1. Cardiovascular inpatients prediction

The first real-world dataset contains the number series of cardiovascular inpatients in major hospitals in Hong Kong and the indices series of air pollutants, *i.e.*, the daily concentrations of nitrogen dioxide (NO2), sulfur dioxide (SO2), ozone (O3), respirable suspended particulate (Rspar), mean daily temperature, relative humidity, etc., which were obtained from air monitoring stations in Hong Kong from 1994 to 1997 [22]. As the previous study has reported the relevance between air pollutants and cardiovascular inpatients [35], the STICM was employed to predict daily cardiovascular disease admissions on the basis of a group of air pollutants (Fig. 3). Thus, for the 14-dimensional system ($n = 14$), the known time points were set as $m = 70$ (days) and the prediction horizon as $L - 1 = 25$ (days). By inferring and selecting the top 11 effective variables, the prediction accuracy of the STICM increases significantly and is better than that of the other methods (Table 1 and Table S1). As shown in Fig. 3i, the STICM uncovers the causal relationship between the admissions of cardiovascular diseases and the air pollutants, in accordance with the literatures [36–38]. The inferred causal relations among the air pollutants also agree with the chemical reactions (Table S4).

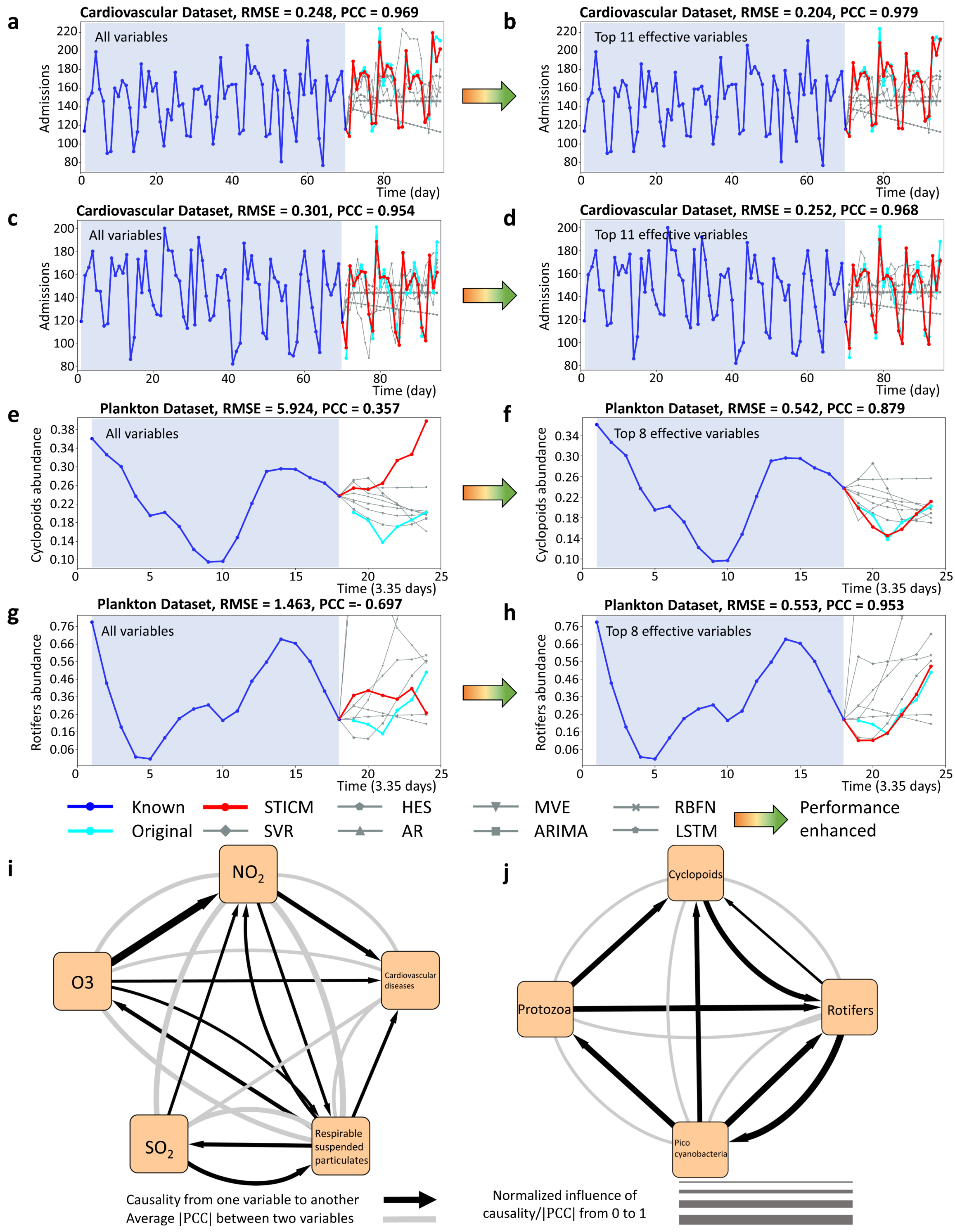

Fig. 3 Future state prediction of cardiovascular admission and plankton abundance. For two periods (a)-(b) and (c)-(d), the STICM predicted the number of cardiovascular admissions based on the high-dimensional time series of air pollutant indices with known length $m = 70$ and prediction horizon $L - 1 = 25$. For two target planktons, *i.e.*, *Cyclopoids* and *Rotifers*, the STICM predicted the dynamic change of their abundance based on the high-dimensional plankton dataset with known length $m = 18$ and prediction horizon $L - 1 = 6$. By selecting the top

11 and top 8 effective variables for the cardiovascular admission dataset and plankton abundance dataset, respectively, the prediction accuracy of the STICM increases significantly ((b), (d), (f), and (h)). The performances of the STICM and other methods are compared in (a)-(h). Based on the STICM, causal networks (i) and (j) were constructed to show the regulatory relationship among cardiovascular admission and air pollutants and that among the plankton, respectively.

### 3.3.2. Plankton density prediction

The STICM was then applied to a dataset collected in a long-term experiment with a marine plankton community isolated from the Baltic Sea from 1990 to 1997 [26,27,39], including the species abundance time series of herbivorous and predatory zooplankton species, several phytoplankton species, detritivores, and bacteria. These plankton species constructed a complex food web. As shown in Fig. 3e-3h, the STICM predicts the dynamic trend of the abundances of two target species (*Cyclopoids* and *Rotifers*), with parameter settings $n = 12$ (total 12 plankton species), $m = 18$ (the known abundance information of 18 steps), and $L - 1 = 6$ (6 step-ahead prediction). By selecting the top 8 effective variables, the STICM achieves a higher prediction accuracy, *i.e.*, $\mathrm{RMSE} = 0.542$ and $\mathrm{PCC} = 0.879$ for cyclopoids and $\mathrm{RMSE} = 0.553$ and $\mathrm{PCC} = 0.953$ for *Rotifers*, than other methods (Table 1 and Table S1). In addition, Fig. 3j depicts the inferred causal network among four species, *i.e.*, *Rotifers*, *Cyclopoids*, *Pico cyanobacteria*, and *Protozoa*. Being consistent with the original food chain network, the causal network also contains other relations among these four species. For example, the links from *Cyclopoids* to *Rotifers* and from *Rotifers* to *Pico cyanobacteria* reveal the fact that the abundance of predators can influence that of the preys. The link from *Protozoa* to *Rotifers* reveals the competitive relation when they have the common predators and preys.

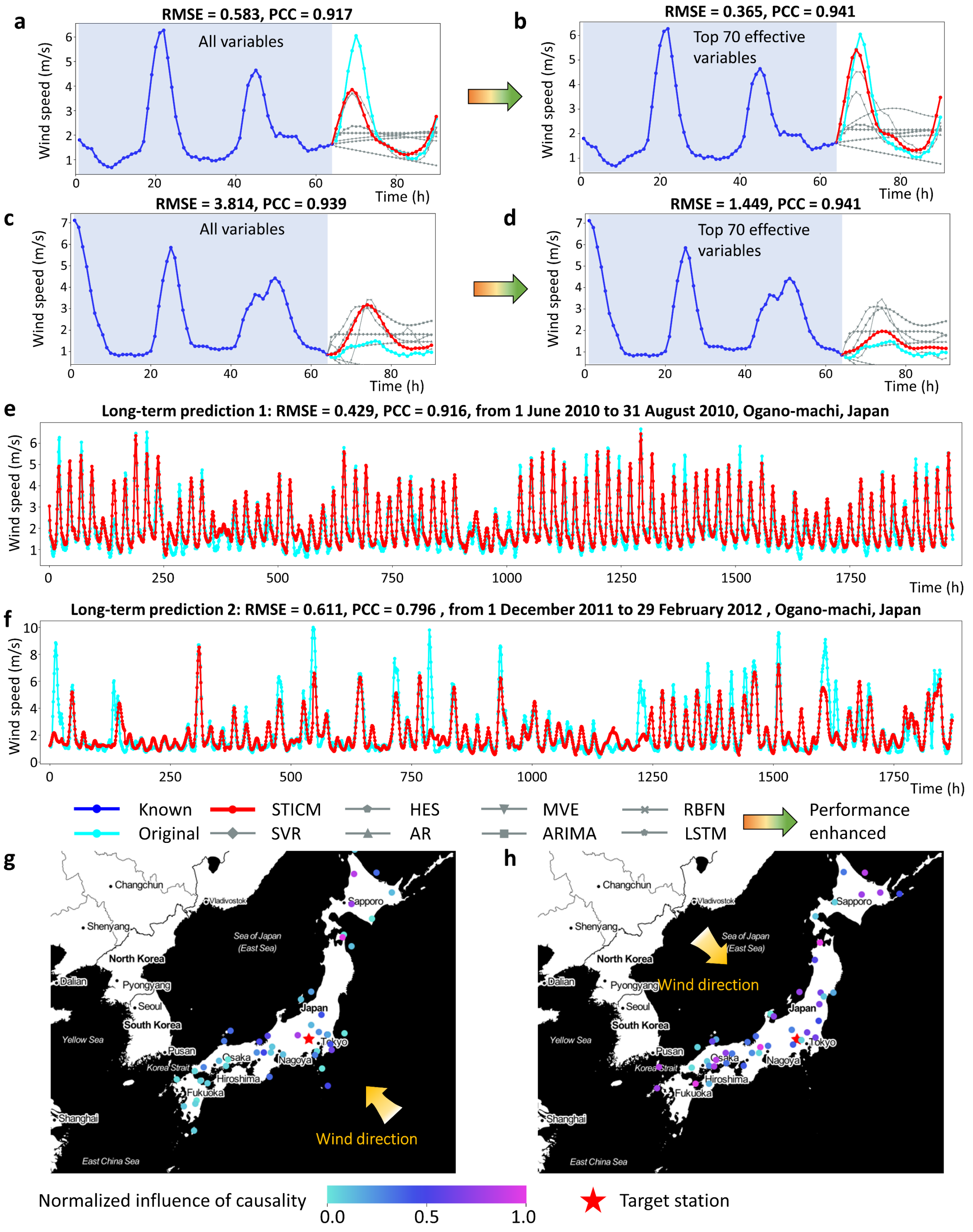

Fig. 4 Wind speed prediction. The STICM predicts the wind speed of a target station around Tokyo marked by a pink star symbol. Based on the time series from all 155 variables (the wind speed of 155 stations) and from the selected top 70 effective variables, the STICM predicted the future wind speed for two periods ((a) and (c) based on all variables and (b) and (d) based on the top 70 variables) with known length $m = 64$ and prediction horizon $L - 1 = 26$. The long-term predictions were performed by the STICM as in (e) and (f), which showed the robustness of the proposed method by predicting the whole season (3 months). The causal relations among the target station and its top 50 effective/causal stations are provided in (g) (for a wet monsoon with wind direction mainly from the southeast) and (h) (for a dry monsoon with wind direction mainly from the northwest).

#### 3.3.3. Wind speed prediction

Wind speed is one of the weather variables with highly time-varying characteristics in nonlinear meteorological systems and is thus extremely difficult to predict. The wind speed dataset was collected from the Japan Methodological Agency [24]. Among the 155 wind stations distributed all around Japan, we selected one target station near Tokyo. As shown in Fig. 4, the STICM predicted the dynamics of the wind speed in the target station with parameter settings $n = 155$, $m = 64$, and $L - 1 = 26$ (Fig. 4a and 4c). After inferring and selecting the 70 most effective variables, the prediction accuracy of the STICM increases significantly, as shown by the comparisons in Fig. 4b and 4d. Based on the effective variables, the predictions of the STICM are better than those of the other methods (Table 1 and Table S1). Long-term predictions were also performed by selecting 70 top effective variables and are provided in Fig. 4e and 4f, from which the wind speed in the target station was continuously predicted for a whole season (3 months). The predictions for more periods are provided in Fig. S5. The inferred causal relations between the locations of top 50 effective variables and that of the target station are consistent with the corresponding monsoon-specific wind directions (Fig. 4g and 4h).

#### 3.3.4. Traffic speed prediction

The transportation system consisting of vehicles, roads and other transportation elements, can be considered as a high-dimensional complex system [40]. Meanwhile, intelligent inspection on such a system is of great importance to city management and development. However, due to the complexity in traffic dynamical systems, predicting the traffic flow precisely is full of challenges. STICM was applied to predict the traffic speed (mile/h), which was based on a dataset generated from $n = 207$ loop detectors in the 134-highway of California, USA. The traffic speed was recorded every five minutes from Mar 1$^{st}$, 2012 to Jun 30$^{th}$, 2012 [29]. In such a dynamic system, each loop detector was considered as a variable by which the traffic speed detected was mainly determined by the observed values from the nearest neighbor sensors. We selected four target sensors, which are the intersections of main roads (Target 1 is located at the intersection of San Diego Freeway and Ventura Freeway; Target 2 is located at the intersection of Hollywood Freeway and Ventura Freeway; Target 3 is located at the intersection of Glendale Freeway and Ventura Freeway; Target 4 is located at the intersection of Hollywood Freeway and Harbor Freeway). Consequently, 55 nearest-neighbor detectors of the target detector were selected to constitute a subsystem. By applying the STICM, the multistep predictions ($L-1=19$ time points ahead) of four target locations/sensors were obtained based on the neighbor 55 variables ($n=55$, Fig. 5a, 5c, 5e, and 5g) and top 30 effective variables ($n=30$, Fig. 5b, 5d, 5f, and 5h) with $m=60$ time points. Based on the effective variables, the RMSEs of the predicted traffic speed on 19 time points significantly decreased, *i.e.*, from 1.757 (Fig. 5a) to 0.852 (Fig. 5b) for Target 1, from 1.551 (Fig. 5c) to 0.536 (Fig. 5d) for Target 2, from 2.207 (Fig. 5e) to 0.762 (Fig. 5f) for Target 3, and from 1.844 (Fig. 5g) to 0.489 (Fig. 5h) for Target 4. The predictions of the STICM are better than those of the other seven prediction methods (Table 1 and Table S1). Supplementary Movie S1 shows the dynamic change in the predicted traffic speed. As shown in Fig. 5i and 5j, most of the causal/effective detectors are distributed around each target detector.

#### 3.3.5. Japan Covid-19 transmission prediction

The pandemic of coronavirus disease 2019 (COVID-19) has posed a global threat to public health. To assist public health departments with their strategic planning, it is important to predict the spread of this infectious disease. The STICM provides a data-driven approach to predict the dynamic change in daily new cases of infectious disease. As shown in Fig. 6, the STICM predicted the number of COVID patients in several cities with severe epidemics in Japan [28], with parameter settings $m=30$ and $L-1=14$. Based on all 47 districts ($n=47$), the predictions of COVID-19 new cases of the six target districts are provided in Fig. 6a (Tokyo), 6c (Tochigi), and 6e (Gunma). After inferring and selecting the top 20 effective/causal districts in each target district, the STICM was predicted much more accurately than the other methods for the six districts (Fig. 6b (Tokyo), 6d (Tochigi), and 6f (Gunma)) The quantitative comparisons were provided in Table 1 and Table S1. Fig. 6g presents the network of COVID-19 transmission in the Kanto region, Japan. The predictions for more districts are provided in Fig. S3.

#### 3.3.6. Meteorological data prediction

The last real-world dataset contains 72-dimensional ground meteorological data ($n=72$) recorded per month in an area around Houston, Galveston, and Brazoria [25] from 1998 to 2004. As shown in Fig. S2, the relative humidity and geopotential height were accurately predicted. For each target index, the STICM was applied to make a 17-step-ahead prediction ($L-1=17$) based on the former $m = 50$ steps of the 72-dimensional data. The prediction results of the STICM are better than those predicted by other seven methods (Table 1 and Table S1).

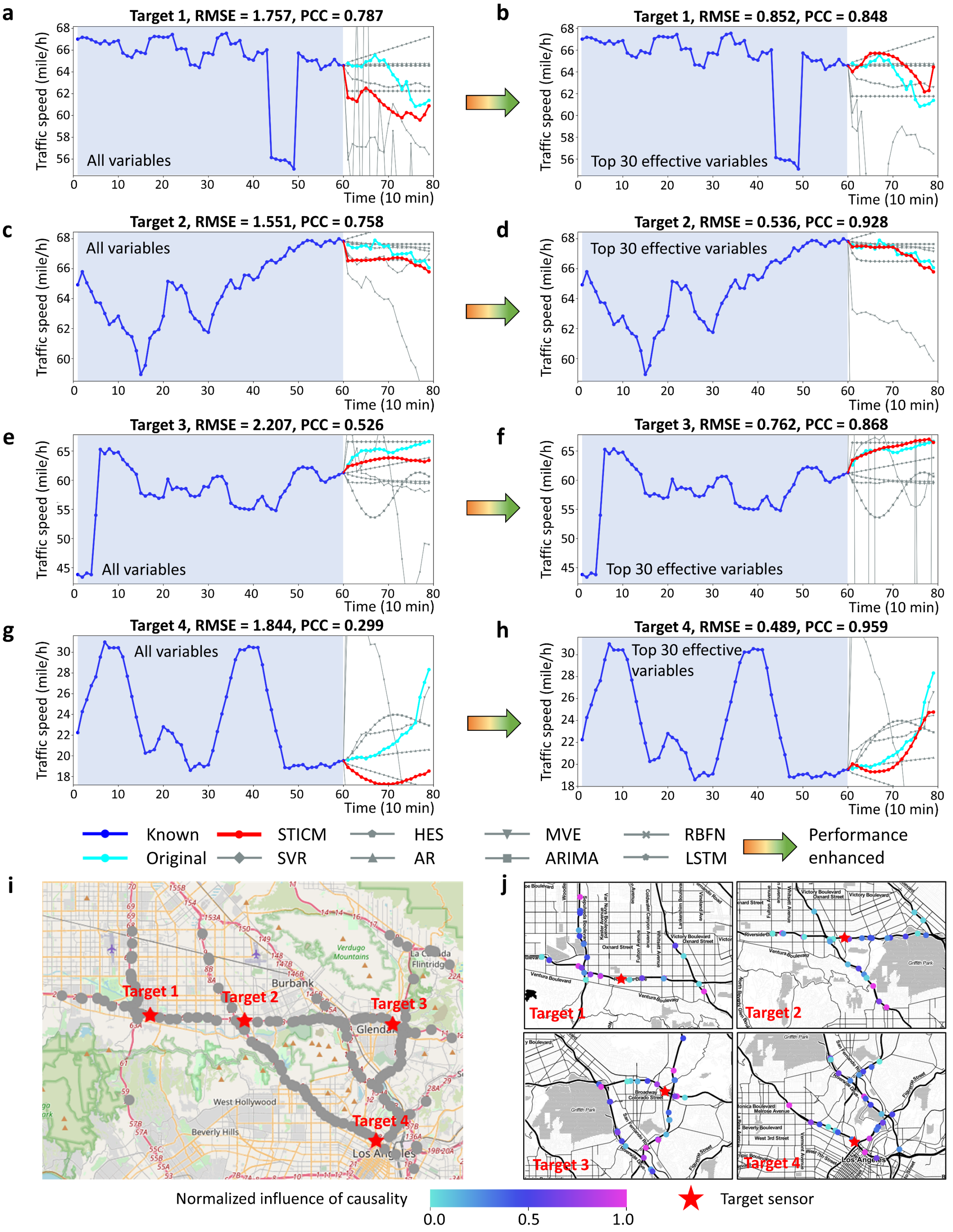

Fig. 5 Traffic speed prediction. Based on the 207-dimensional traffic speed dataset, the STICM predicted the traffic speed of four target locations/sensors with 60-step known information ($m = 60$) and 19-step prediction horizon ($L - 1 = 19$), *i.e.*, (a) and (b) for target 1, (c) and (d) for target 2, (e) and (f) for target 3, (g) and (h) for target 4, where the four target locations were marked by red star symbols in (i). By inferring and selecting the top

30 effective variables (*i.e.*, the effective traffic speeds in 30 locations), the prediction accuracy of the STICM significantly increases and is better than that of the other methods ((b), (d), (f), and (h)). The associations/causal relations among the neighboring locations/sensors are shown in (j).

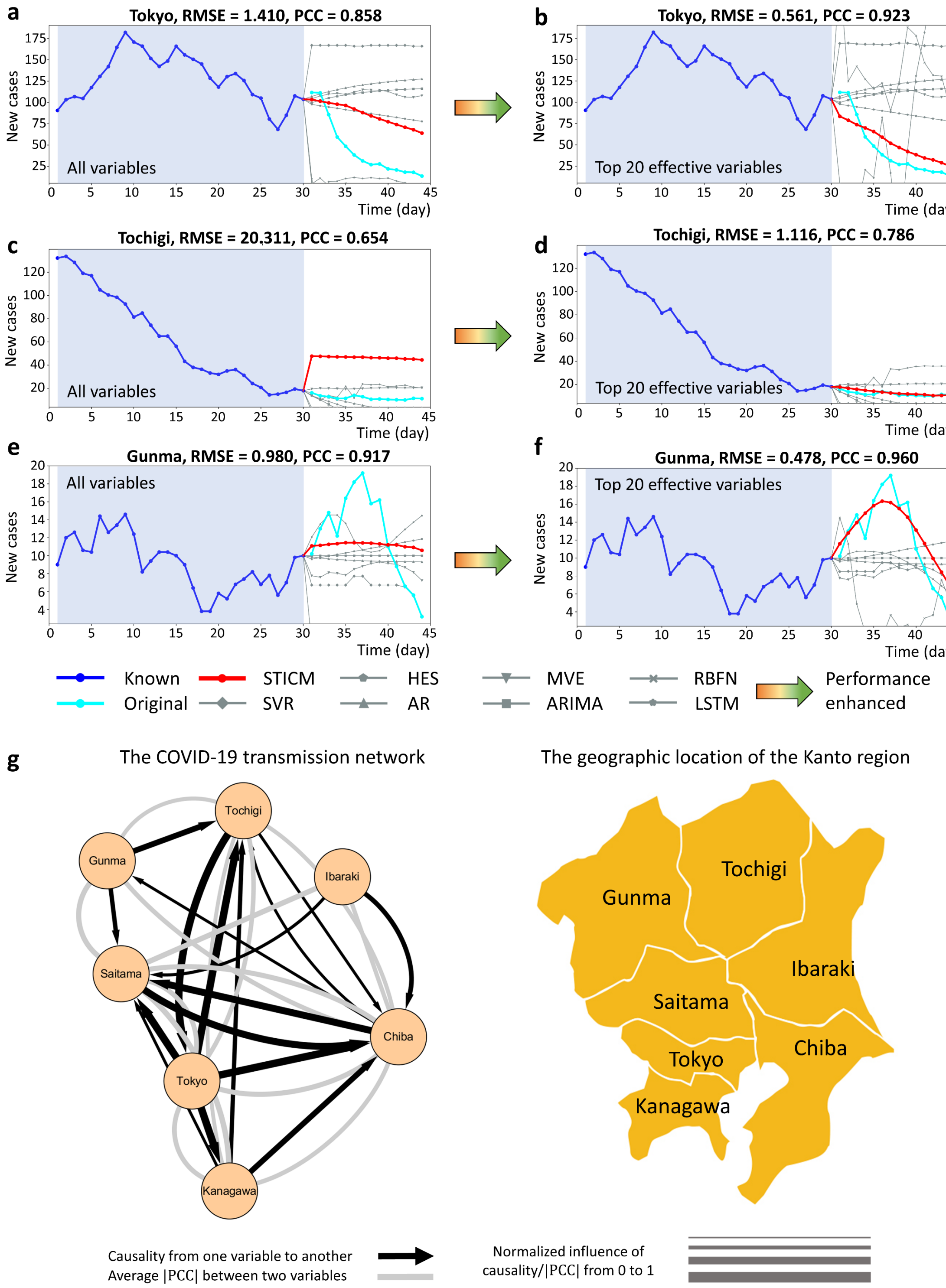

Fig. 6 Predicting the number of COVID-19 patients. Based on the time series of COVID-19 new cases of 47 districts (the left subfigures) or selected top 20 effective districts in each target district (the right subfigures), the STICM predicts the numbers of future new cases, with 30-step known information ($m = 30$) and 14-step prediction horizon ($L - 1 = 14$), *i.e.*, (a) and (b) for Tokyo, (c) and (d) for Tochigi, (e) and (f) for Gunma. Based

on the STICM, the casual network (g) of COVID-19 transmission in the Kanto region, Japan revealed the regulatory relationship in terms of COVID-19 spread among the districts in this region.

## 4. Discussion

In this work, we proposed the STICM framework to achieve the multistep-ahead prediction with causal factor inference based on high-dimensional time series in a robust way. Through STICM, the spatiotemporal information of high-dimensional observables is transformed to the temporal information of a target variable on the basis of the delay embedding theory. That is, the primary STI form is an encoder which transforms the spatiotemporal matrix $[\mathbf{X}^{t-w}, \mathbf{X}^{t-w+1}, \dots, \mathbf{X}^{t}]$ to the temporal vector $\mathbf{Y}^t$ of a target variable by $F$, while the conjugate STI form recovers the temporal vector $\mathbf{Y}^t$ back to the original matrix $[\mathbf{X}^{t-w}, \mathbf{X}^{t-w+1}, \dots, \mathbf{X}^{t}]$ by $F^{-1}$. Training $F$ and $F^{-1}$ simultaneously in a semi-supervised manner, the STICM solves the STI equations and makes the prediction highly robust, as shown in the applications. Clearly, the multiple future/unknown values $\{y^{m+1}, y^{m+2}, \dots, y^{m+L-1}\}$ are obtained concurrently by the STICM, indicating that the proposed method makes the multistep-ahead prediction. Moreover, the STICM carries out causal inference based on Granger causality and thus identifies the causal/effective variables on the target variable. Causal inference enables a deep understanding of the intrinsic dynamics of the complex system, thus providing the interpretability of the STICM, and to a considerable extent, reduces the dimension. Thus, the prediction accuracy is enhanced by selecting the effective variables for prediction. A series of applications show that the STICM achieves better performance than seven traditional prediction approaches. However, there are limitations of Granger causality that it fails to reveal the real causality in some cases. In the future, we will explore the causality relationship in different perspectives for better investigating the intrinsic dynamics of a complex system.

In conclusion, the proposed STICM framework has the following advantages compared with traditional prediction methods. First, the STICM is capable of exploring the time-series data and transforming the spatial information of high-dimensional observables into the temporal information of a target variable. Second, once being trained in a semi-supervised manner, the STICM well solves the primary and conjugate STI equations simultaneously (corresponding to a spatiotemporal convolutional autoencoder), thus making the multistep-ahead predictions robustly even in noise-perturbed cases. Third, in practical applications, the STICM can distinguish the effective/relevant variables, thus unveiling the underlying causal mechanism (in the sense of Granger causality) among massive observables of the dynamical systems. In addition, building on a solid theoretical background of the STI equations and with the TCN causal convolution structure, the STICM opens a new way to explore the spatiotemporal information from high-dimensional time series, and has been validated by the applications to a variety of real-world scenarios.

## 5. Acknowledgements

We thank the Japan Meteorological Agency which provided the datasets of wind speeds used in this study (available via the Japan Meteorological Business Support Center). This work was supported by the National Key R&D Program of China (No. 2017YFA0505500), National Natural Science Foundation of China (Nos. 11771152, 11901203, 11971176, 31930022, and 31771476), and Japan Science and Technology Agency Moonshot R&D (No. JPMJMS2021).

## 6. Data availability

All data needed to validate the conclusions are present in the paper and/or the Supplementary Materials. All data are available at https://github.com/mahp-scut/STICM.

## 7. Code availability

The code used in this study is available at https://github.com/mahp-scut/STICM.

## 8. Reference

[1] Box GE, Pierce DA. Distribution of residual autocorrelations in autoregressive-integrated moving average time series models. Journal of the American Statistical Association 1970;65:1509–26.
[2] Rousseeuw PJ, Leroy AM. Robust regression and outlier detection. vol. 589. John wiley & sons; 2005.
[3] Holt CC. Forecasting seasonals and trends by exponentially weighted moving averages. International Journal of Forecasting 2004;20:5–10.
[4] Brown RG. Exponential smoothing for predicting demand. Operations Research, vol. 5, INST OPERATIONS RESEARCH MANAGEMENT SCIENCES 901 ELKRIDGE LANDING RD, STE 400, LINTHICUM HTS, MD 21090-2909; 1957, p. 145–145.
[5] Karevan Z, Suykens JA. Transductive LSTM for time-series prediction: An application to weather forecasting. Neural Networks 2020.

[6] Hochreiter S, Schmidhuber J. Long short-term memory. Neural Computation 1997;9:1735–80.
[7] Connor JT, Martin RD, Atlas LE. Recurrent neural networks and robust time series prediction. IEEE Transactions on Neural Networks 1994;5:240–54.
[8] Wei WW. Time series analysis. The Oxford Handbook of Quantitative Methods in Psychology: Vol. 2, 2006.
[9] Wang W-X, Lai Y-C, Grebogi C. Data based identification and prediction of nonlinear and complex dynamical systems. Physics Reports 2016;644:1–76.
[10] Weigend AS. Time series prediction: forecasting the future and understanding the past. Routledge; 2018.
[11] Chen P, Liu R, Aihara K, Chen L. Autoreservoir computing for multistep ahead prediction based on the spatiotemporal information transformation. Nat Commun 2020;11:4568. https://doi.org/10.1038/s41467-020-18381-0.
[12] Ma H, Leng S, Aihara K, Lin W, Chen L. Randomly distributed embedding making short-term high-dimensional data predictable. Proc Natl Acad Sci USA 2018;115:E9994–10002. https://doi.org/10.1073/pnas.1802987115.
[13] Sauer T, Yorke JA, Casdagli M. Embedology. Journal of Statistical Physics 1991;65:579–616.
[14] Takens F. Detecting strange attractors in turbulence. Dynamical systems and turbulence, Warwick 1980, Springer; 1981, p. 366–81.
[15] Casdagli M. Nonlinear prediction of chaotic time series. Physica D: Nonlinear Phenomena 1989;35:335–56.
[16] Bai S, Kolter JZ, Koltun V. An Empirical Evaluation of Generic Convolutional and Recurrent Networks for Sequence Modeling. ArXiv:180301271 [Cs] 2018.
[17] Gehring J, Auli M, Grangier D, Dauphin YN. A convolutional encoder model for neural machine translation. ArXiv Preprint ArXiv:161102344 2016.
[18] Lea C, Flynn MD, Vidal R, Reiter A, Hager GD. Temporal Convolutional Networks for Action Segmentation and Detection. ArXiv:161105267 [Cs] 2016.
[19] Dauphin YN, Fan A, Auli M, Grangier D. Language modeling with gated convolutional networks. International conference on machine learning, PMLR; 2017, p. 933–41.
[20] Cho K, Van Merriënboer B, Gulcehre C, Bahdanau D, Bougares F, Schwenk H, et al. Learning phrase representations using RNN encoder-decoder for statistical machine translation. ArXiv Preprint ArXiv:14061078 2014.
[21] Curry JH. A generalized Lorenz system. Communications in Mathematical Physics 1978;60:193–204.
[22] Wong TW, Lau TS, Yu TS, Neller A, Wong SL, Tam W, et al. Air pollution and hospital admissions for respiratory and cardiovascular diseases in Hong Kong. Occupational and Environmental Medicine 1999;56:679–83.
[23] Fan J, Zhang W, others. Statistical estimation in varying coefficient models. The Annals of Statistics 1999;27:1491–518.
[24] Hirata Y, Aihara K. Predicting ramps by integrating different sorts of information. The European Physical Journal Special Topics 2016;225:513–25.
[25] Zhang K, Fan W. Forecasting skewed biased stochastic ozone days: analyses, solutions and beyond. Knowl Inf Syst 2008;14:299–326. https://doi.org/10.1007/s10115-007-0095-1.

[26] Beninca E, Huisman J, Heerkloss R, Johnk KD, Branco P, Scheffer M, et al. Chaos in a long-term experiment with a plankton community 2008;451:5.

[27] Benincà E, Jöhnk KD, Heerkloss R, Huisman J. Coupled predator-prey oscillations in a chaotic food web: Coupled predator-prey oscillations. Ecology Letters 2009;12:1367–78. https://doi.org/10.1111/j.1461-0248.2009.01391.x.

[28] Wahltinez O, others. COVID-19 Open-Data: curating a fine-grained, global-scale data repository for SARS-CoV-2 2020.

[29] Li Y, Yu R, Shahabi C, Liu Y. Diffusion Convolutional Recurrent Neural Network: Data-Driven Traffic Forecasting. ArXiv:170701926 [Cs, Stat] 2018.

[30] Thombs LA, Schucany WR. Bootstrap prediction intervals for autoregression. Journal of the American Statistical Association 1990;85:486–92.

[31] Howlett RJ, Jain LC. Radial basis function networks 2: new advances in design. vol. 67. Physica; 2013.

[32] Ye H, Sugihara G. Information leverage in interconnected ecosystems: Overcoming the curse of dimensionality. Science 2016;353:922–5. https://doi.org/10.1126/science.aag0863.

[33] Wang L. Support vector machines: theory and applications. vol. 177. Springer Science & Business Media; 2005.

[34] Tong H, Ng MK. Calibration of $\epsilon$- insensitive loss in support vector machines regression. Journal of the Franklin Institute 2019;356:2111–29.

[35] Xia Y, Härdle W. Semi-parametric estimation of partially linear single-index models. Journal of Multivariate Analysis 2006;97:1162–84.

[36] Devlin RB, Duncan KE, Jardim M, Schmitt MT, Rappold AG, Diaz-Sanchez D. Controlled exposure of healthy young volunteers to ozone causes cardiovascular effects. Circulation 2012;126:104–11.

[37] Lee B-J, Kim B, Lee K. Air pollution exposure and cardiovascular disease. Toxicological Research 2014;30:71–5.

[38] Luo K, Li R, Li W, Wang Z, Ma X, Zhang R, et al. Acute effects of nitrogen dioxide on cardiovascular mortality in Beijing: an exploration of spatial heterogeneity and the district-specific predictors. Scientific Reports 2016;6:1–13.

[39] Heerkloss R, Klinkenberg G. A long-term series of a planktonic foodweb: a case of chaotic dynamics. Internationale Vereinigung Für Theoretische Und Angewandte Limnologie: Verhandlungen 1998;26:1952–6.

[40] Wang J, Lv W, Jiang Y, Qin S, Li J. A multi-agent based cellular automata model for intersection traffic control simulation. Physica A: Statistical Mechanics and Its Applications 2021;584:126356.

**Table 1 Comparison of the performance among eight prediction methods.**

| **Dataset** | **Metric**[a] | **Methods** | | | | | | | |
|---|---|---|---|---|---|---|---|---|---|
| | | **STICM** | **MVE** | **AR** | **ARIMA** | **HES** | **LSTM** | **RBFN** | **SVR** |
| Lorenz System (noise-free) | RMSE | **0.111** | 1.498 | 1.546 | 1.685 | 1.564 | 0.806 | 1.798 | 2.024 |
| | PCC | **0.997** | 0.731 | -0.66 | 0.297 | -0.637 | 0.992 | -0.419 | 0.193 |
| Lorenz system with noise ($\sigma = 0.5$) | RMSE | **0.307** | 1.607 | 1.486 | 1.382 | 1.565 | 1.62 | 1.86 | 2.026 |
| | PCC | **0.989** | 0.711 | -0.66 | -0.339 | -0.644 | -0.15 | 0.29 | 0.218 |
| Cardiovascular inpatients | RMSE | **0.228** | 0.968 | 1.071 | 1.065 | 1.391 | 1.104 | 0.994 | 0.804 |
| | PCC | **0.974** | 0.467 | 0.351 | 0.366 | -0.157 | 0.21 | 0.244 | 0.865 |
| Plankton density | RMSE | **0.548** | 1.669 | 1.441 | 0.776 | 2.408 | 3.647 | 3.728 | 2.84 |
| | PCC | **0.917** | 0.522 | 0.359 | 0.781 | -0.372 | 0.377 | -0.503 | 0.412 |
| Wind speed | RMSE | **0.908** | 2.632 | 1.348 | 3.28 | 5.144 | 2.243 | 1.985 | 2.384 |
| | PCC | **0.942** | 0.895 | -0.28 | 0.817 | 0.417 | -0.189 | 0.873 | 0.321 |
| Traffic speed | RMSE | **0.66** | 2.248 | 2.344 | 3.135 | 2.728 | 4.597 | 6.676 | 3.544 |
| | PCC | **0.901** | 0.359 | 0.044 | 0.162 | -0.434 | 0.204 | -0.181 | 0.265 |
| Japan Covid-19 transmission | RMSE | **0.608** | 2.311 | 2.553 | 4.148 | 2.819 | 4.031 | 3.48 | 6.16 |
| | PCC | **0.9** | 0.012 | 0.049 | -0.037 | 0.356 | 0.016 | 0.405 | 0.422 |
| Meteorological data | RMSE | **0.811** | 0.935 | 1.065 | 1.029 | 1.267 | 1.154 | 1.278 | 1.165 |
| | PCC | **0.728** | 0.324 | 0.015 | 0.093 | -0.171 | 0.067 | -0.053 | 0.341 |

[a]The performance metrics include the values of the root-mean-square error (RMSE) and the Pearson correlation coefficient (PCC). The RMSE was normalized by the standard deviation of the real data.